\pdfoutput=1

\documentclass[11pt]{article}

\usepackage{ACL2023}

\usepackage{times}
\usepackage{latexsym}

\usepackage[T1]{fontenc}

\usepackage[utf8]{inputenc}

\usepackage{microtype}

\usepackage{inconsolata}

\usepackage{multirow}
\usepackage{booktabs}
\usepackage{caption}
\usepackage{subcaption}
\usepackage{xspace}
\usepackage{amsmath}
\usepackage{amssymb}
\usepackage{newtxmath}
\usepackage{bbm}
\usepackage{graphicx}
\usepackage{tabularx}
\usepackage{bbding}
\usepackage{pifont}
\usepackage{url}
\usepackage[ruled,vlined]{algorithm2e}
\usepackage{diagbox}
\usepackage{makecell}
\usepackage{colortbl}
\usepackage{bbm}
\usepackage{color}
\usepackage{xcolor}
\usepackage{hhline}
\usepackage{soul}

\newcommand{\paratitle}[1]{\vspace{1.5ex}\noindent\textbf{#1}}
\newcommand{\ie}{\emph{i.e.,}\xspace}

\newcommand{\eg}{\emph{e.g.,}\xspace}

\newcommand{\ignore}[1]{}

\title{Improving Large Language Models via Fine-grained Reinforcement Learning with Minimum Editing Constraint}

\author{
    \textbf{Zhipeng Chen\textsuperscript{{1},{3}}\thanks{\llap{}\:\:\:Equal contribution. },
	        Kun Zhou\textsuperscript{{2},{3}}\footnotemark[1],
	        Wayne Xin Zhao\textsuperscript{{1},{3}}\thanks{\llap{}\:\:\:Corresponding author. },
                Junchen Wan\textsuperscript{{4}}}, \\
                \textbf{Fuzheng Zhang\textsuperscript{{4}},
                Di Zhang\textsuperscript{{4}}~\and
	        Ji-Rong Wen\textsuperscript{{1},{2},{3}}}\\
	\textsuperscript{1}Gaoling School of Artificial Intelligence, Renmin University of China.\\
	\textsuperscript{2}School of Information, Renmin University of China.\\
	\textsuperscript{3}Beijing Key Laboratory of Big Data Management and Analysis Methods.\\
        \textsuperscript{4}Kuaishou Technology, Beijing, China.\\
    	\texttt{zhipeng\_chen@ruc.edu.cn,francis\_kun\_zhou@163.com,batmanfly@gmail.com}
}

\begin{document}
\maketitle
\begin{abstract}
Reinforcement learning (RL) has been widely used in training large language models~(LLMs) for preventing unexpected outputs, \eg reducing harmfulness and errors.
However, existing RL methods mainly adopt instance-level reward, which cannot provide fine-grained supervision for complex reasoning tasks. As a result, the RL training cannot be fully aware of the specific part or step that actually leads to the incorrectness in model response.
To address it, we propose a new RL method named \textbf{RLMEC} that incorporates a generative model as the reward model, which is trained by the erroneous solution rewriting task under the minimum editing constraint, which can produce token-level supervision for RL training.
Based on the generative reward model, we design the token-level RL objective for training and an imitation-based regularization for stabilizing RL process. And these two objectives focus on the revision of the key tokens for the erroneous solution, reducing the effect of other unimportant tokens.
Experiment results on 8 tasks have demonstrated the effectiveness of our approach.
Our code and data are available at \url{https://github.com/RUCAIBox/RLMEC}.

\end{abstract}

\section{Introduction}

\definecolor{intro_neg}{RGB}{254, 179, 189}
\definecolor{intro_pos}{RGB}{221, 241, 213}

\begin{figure}[t]
    \centering
    \includegraphics[width=0.49\textwidth]{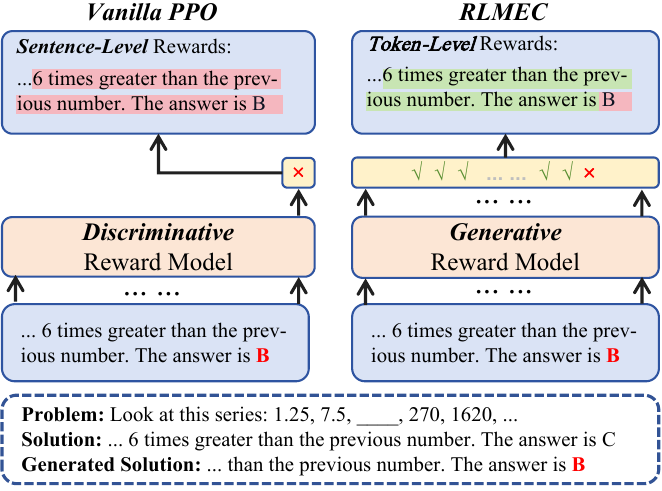}
    \caption{The comparison of our generative reward model and the traditional discriminative one in PPO. Red and green background colors denote \colorbox{intro_neg}{negative} and \colorbox{intro_pos}{positive} rewards, respectively.}
    \label{rewriting_model}
\end{figure}

Owing to unsupervised pre-training on large-scale text corpora, large language models (LLMs) have shown remarkable performance on various text generation tasks~\cite{LLMSurvey,PaLM-2}, such as question answering, summarization and translation~\cite{GPT-4}.
To further improve the task solving capacity,  researchers~\cite{llama2,qwen} have proposed supervised fine-tuning (SFT) and reinforcement learning (RL) methods, which can better adapt LLMs to specific domains or downstream tasks after pre-training.
Typically, SFT methods~\cite{instructgpt,flan_collection} incorporate annotated input-output pairs (\eg question and solution, instruction and response) to train the LLM for learning the sequence-to-sequence pattern;  
RL methods~\cite{ppo,Christiano-NeurIPS-2017-Deep} adopt a reward model to measure the quality of the generated outputs from the LLM, and then guide its training for maximizing and minimizing the expectation of generating high-quality and low-quality ones, respectively.

As RL methods are capable of directly reducing the probability of  LLMs for producing unexpected outputs, they have been widely used in optimizing LLMs towards better human alignment (\eg reducing harmfulness) and  stronger ability (\eg reducing errors~\cite{wizardmath,math-shepherd}).
Generally, RL methods first train a discrimination model for distinguishing desirable and undesirable outputs.
Then, the model is used to produce the reward scores for the sampled outputs from the LLM, and the LLM would be trained by encouraging and punishing the generation of high-score and low-score ones accordingly.


Despite the success, as existing RL methods mostly utilize instance-level reward for each sampled output, it is often difficult to provide accurate fine-grained supervision on complex reasoning tasks (\eg mathematical reasoning).
Concretely, given a complex task, the sampled outputs from the LLM tend to be highly similar in surface expression, only with key differences in few specific words or steps~\cite{rft} that determine the correctness. 
We argue that instance-level RL approaches~\cite{instructgpt,Christiano-NeurIPS-2017-Deep,moss_rlhf} have two major limitations. 
First, as the unimportant parts would often occupy a large amount of supervision signals,  instance-level rewards can not accurately emphasize the more important evidence related to correctness, leading to inefficient or redundant supervision. 
Second, the paired correct and incorrect outputs may share the overlapping content but receive opposite optimization goals, which may lead to the optimization conflict issue on such overlapped content, 
making it still infeasible to provide 
accurate fine-grained supervisions. 

To address these issues, in this paper, we propose a novel method, \emph{Reinforcement Learning with Minimum Editing Constraint}~(\textbf{RLMEC}), to improve the training of LLMs by fine-grained supervision signals.  
Our approach is inspired by the homework correction process of professional teachers, in which she/he first identifies the incorrect parts and then provides necessary revisions or comments accordingly. Following such an idea, we train a generative reward model by an erroneous solution rewriting task under the constraint of minimum editing distance. 
The reward model plays a similar role to teachers by producing fine-grained supervision, \ie token-level quality assessment scores.  
Instead of using a new demonstration as positive, our reward model tries to correct the output with minimum edits. 
Specially, we utilize the specially trained reward model to produce the token probabilities for computing the token-level rewards, and optimize the LLM using the proximal policy optimization method~(PPO)~\cite{ppo}. 
In this way, by contrasting the original and corrected outputs, 
the LLM would be instructed more informatively, thus becoming aware of the correct way to generate the response.  
Figure~\ref{rewriting_model} illustrates the comparison of the Vanilla PPO and our proposed RLMEC approach. 

The major novelty of this paper lies in the incorporation of a generative reward models with minimum editing constraint for RL training of LLMs.
Table~\ref{comparision} presents the major differences between our method and previous work. 
To evaluate the effectiveness of our methods, we conduct the experiment on two types of complex reasoning tasks, \ie question answering~\cite{ecqa,obqa} and mathematical reasoning~\cite{gsm8k,math}.
In these evaluation tasks, our RLMEC mostly outperforms other competitive SFT and RL methods, based on 7B and 13B LLMs.
Moreover, our analysis experiments also show that our method is able to stabilize the RL training process and reduce the erroneous steps in the sampled outputs of LLMs. 
\section{Related Work}

\paratitle{Reinforcement Learning for LLMs.}
With the development of the LLMs, reinforcement learning (RL)~\cite{Christiano-NeurIPS-2017-Deep, rlhf2} is widely utilized to further improve the ability of LLMs.
Proximal Policy Optimization (PPO)~\cite{ppo} is the traditional algorithm to employ RL.
To provide fine-grained supervision signals, previous work~\cite{ppo_a2c,moss_rlhf} utilizes the critic model to calculate the reward of the current stage.
Because of the instability of the training procedure of reinforcement learning, recent work~\cite{dpo,coh,quark,slichf} has utilized supervised-finetuning (SFT) to simulate the RL procedure.
These methods fuse the quality of the responses into the supervision signals.
Moreover, existing work~\cite{Uesato-2022-Solving,wizardmath,math-shepherd,aft,preference-grounded-guidance} has found that process-supervision signals can better guide the training process of LLMs.
Besides, other methods~\cite{spo, spin} improve the ability of LLMs during self-play procedure.
In this work, we proposed a new RL framework with generative reward model to directly provide the fine-grained supervisions, which enable to focus on few key tokens.

\paratitle{LLMs for Reasoning.}
Previous work utilizes two types of methods (\ie prompting and training) to enhance the reasoning ability of LLMs.
For the prompting methods, Chain-of-Thought (CoT)~\cite{cot,zero-shot-cot} guides LLMs to generate the intermediate reasoning steps before generating the final answer.
Based on CoT, previous work decomposes the problem into several simple sub-problems~\cite{Sccessive-Prompting}, utilizes the external tools to help LLMs~\cite{pal,react,toolformer,chatcot}, designs the specific agents to perform reasoning~\cite{eot,multi-agent}, or post-process the generated response~\cite{self-refine,self-consistency}.
Besides, existing work also guides LLMs to perform reasoning in the specific structure, \eg tree~\cite{tot,xot} or graph~\cite{got}.
For the training methods, previous work~\cite{minerva,zhao2022jiuzhang} has leveraged domain-specific data to fine-tune the LLMs.
Because of the limitation of the training data, the data generated by teacher model (\eg GPT-4) is utilized to augment the training data~\cite{mammoth,metamath,tora,zhao2023jiuzhang,zhou2024jiuzhang30}.
In this work, we aim to train the LLMs via fine-grained RL to improve their reasoning ability.
\section{Preliminary}
\label{preliminary}

In this work, we focus on improving the performance of LLMs on complex reasoning tasks with 
reinforcement learning~(RL) algorithm. 
Typically, complex reasoning tasks require LLMs to perform step-by-step reasoning (\eg chain-of-thought~\cite{cot,zero-shot-cot}) for each question, where LLMs progressively generate the solution for reaching the answer.
In this process, LLMs are prone to make mistakes at the intermediate steps, which likely lead to totally wrong answer~\cite{halu,zhang2023evaluating}. 
Our goal is to optimize a pre-trained LLM using RL algorithm, to reduce its errors and improve the task performance. 

Formally, we are given a collection of question-solution pairs, denoted as $\mathcal{D}=\{\langle q_i,s_i \rangle\}_{i=1}^{n}$, where each question and solution are both composed by a sequence of tokens, denoted as $\{t_{0}, \cdots, t_{m}\}$.
Then, we follow the proximal policy optimization~(PPO) framework~\cite{ppo} for RL, and make improvements about reward model and training loss.
In PPO, the LLM to be optimized is the \emph{policy model}, and its original parameters would be copied to compose the \emph{reference model}. 
During training, the reference model outputs the sampled solutions for the given question, denoted as $\hat{s}$, and then the policy model would learn from the feedback from a \emph{reward model}, which produces the reward $R_{\hat{s}}$ for the sampled output $\hat{s}$. 
Based on it, the parameters of the policy model will be optimized to maximize the reward expectation of all the sampled outputs, and the target function is:
\begin{equation}
\small
\label{ppo_reward}
\mathcal{J}(\theta) = \sum_{i=1}^{n}r(q_i, \hat{s}_i) \times R_{\hat{s}_i}, ~~r(q_i,\hat{s}_i)=\frac{P_\theta(\hat{s}_i|q_i)}{P_{\theta'}(\hat{s}_i|q_i)},    
\end{equation}
where $r(q_i,\hat{s}_i)$ is the coefficient of importance sampling, $\theta$ and $\theta'$ are the parameters of policy model and reference model, respectively.
\section{Approach}

\begin{figure*}[h]
    \centering
    \includegraphics[width=\textwidth]{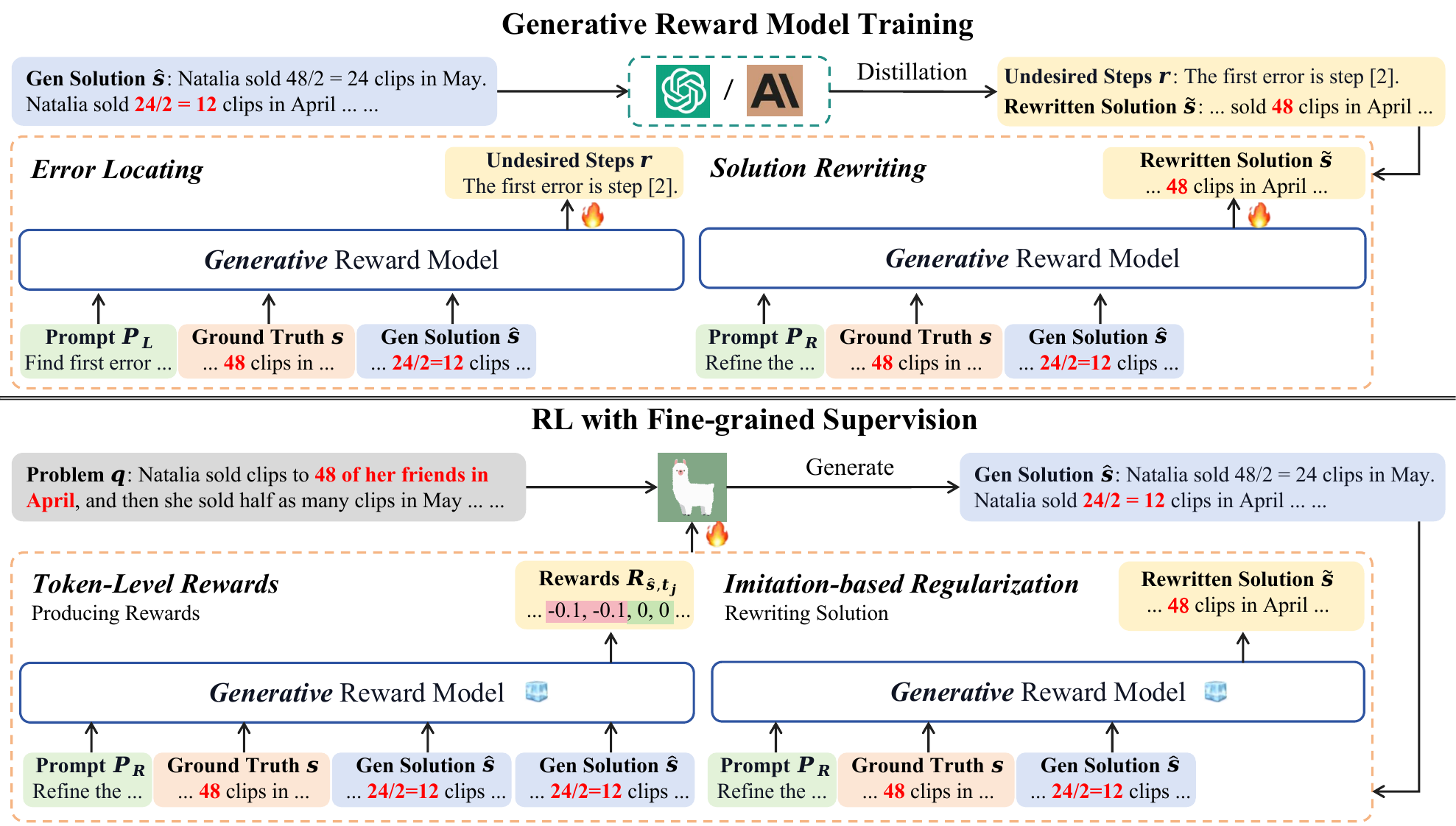}
    \caption{The overview of our RLMEC. Based on the sampled LLM solutions that contain errors, we train the generative reward model using the erroneous solution rewriting task and the distilled data with minimum editing constraint from the teacher model. Then, we perform RL training on the policy model (\ie our LLM) with fine-grained supervision using the token-level RL objective and the imitation-based regularization.}
    \label{framework}
\end{figure*}

In this section, we present our proposed RLMEC, a new RL approach for improving LLMs on complex reasoning tasks.
In RLMEC, we train a generative reward model to produce token-level reward scores for the sampled outputs from the policy model (\ie the LLM), then optimize the policy model via RL based on the fine-grained rewards.
Figure~\ref{framework} illustrates the overall framework of our RLMEC.

\subsection{Generative Reward Model Training}
To provide fine-grained supervision for RL, we train a generative model based on the sequence-to-sequence loss as the reward model.
For a given task, the reward model aims to offer estimations for all the output tokens about their correctness. 
To achieve this, we design an \emph{erroneous solution rewriting} task with the constraint of minimum editing distance to train the reward model, enabling it to focus on the key tokens that lead to the final wrong answer for punishing. 


\paratitle{Erroneous Solution Rewriting.} 
This task aims to correct the error tokens in the LLM generated solutions with minimum edits. 
Formally, given the question $q$, ground-truth solution $s$, and the  generated solution $\hat{s}$, we  rewrite $\hat{s}$ into a correct solution $\Tilde{s}$.
Specifically, we decompose it into two sub-tasks, \ie error locating and solution rewriting.
For error locating, the model requires to locate the first erroneous reasoning step in $\hat{s}$, which would mislead the following steps into erroneous ones. 
Concretely, we split $\hat{s}$ into a sequence of reasoning steps according to the full stop or question mark: $\hat{s}=\{r_{0},r_{1},\dots,r_{n}\}$.
Then, the reward model needs to find the first undesired reasoning step $r_t$ based on the given question and ground-truth solution:
\begin{equation}
\small
RM({p}_{L},q,s,\hat{s})\rightarrow r_t, \label{eq-locate}
\end{equation}
where $p_{L}$ is the prompt to guide the model.
Then, for solution rewriting, we leverage another prompt $p_{R}$ to guide the reward model that rewrites the erroneous steps after $r_t$ in $\hat{s}$ into the correct $\Tilde{s}$:
\begin{equation}
\small
    RM(p_{R},q,s,\hat{s},r_t)\rightarrow \Tilde{s}. \label{eq-rewrite}
\end{equation}
By training on the two tasks, the generative reward model would be able to rewrite erroneous solutions with the minimum editing constraint.


\paratitle{Distillation with Minimum Editing Constraint.}
To train the reward model for fulfilling the above two subtasks, we collect the data from a powerful \emph{teacher LLM}~(\ie Claude 2~\cite{claude2}) to distill the task knowledge for our reward model, while other models (\eg GPT-4) or human annotators can be also applied.  
Concretely, we first sample the generated solutions from our LLM, and select the wrong ones to compose the erroneous solution set $\{\hat{s}\}$.
Then, we feed the given question $q$, ground-truth solution $s$, and the generated erroneous solution $\hat{s}$ into the teacher LLM, and add several annotated exemplars into the prompt, to guide the generation of the first error step $r_t$ and the correct rewritten solution $\Tilde{s}$.
Here, in-context exemplars are human-crafted high-quality instances, and the ones for solution rewriting strictly satisfy the minimum editing constraint with only very few revised tokens.
Therefore, we can obtain high-quality synthetic distilled data for the two subtasks.
Finally, following Eq.~\eqref{eq-locate} and Eq.~\eqref{eq-rewrite}, we prepare the inputs and outputs for the two subtasks, and merge them for training our reward model.

\subsection{RL with Fine-grained Supervision}
After training the generative reward model, we can leverage it to produce fine-grained supervision for the RL training of the policy model (\ie our LLM).
We obtain the token-level rewards based on the generated probabilities from the reward model, and design the token-level RL objective with the imitation-based regularization for training our LLM. 

\paratitle{Token-level Reward  Generation.}
After distillation, the generative reward model can rewrite the original solution to provide the correct one.
Owing to the minimum editing constraint, the error tokens would receive lower probabilities because they should be replaced by other tokens, and the correct tokens would obtain higher probabilities.
Therefore, we can utilize the token probabilities from the reward model to assign the token-level rewards.
This is quite distinct from the conventional reward model~\cite{instructgpt} which only produces instance-level reward scores. 
Concretely, given the prompt $p_{R}$, question $q$, ground-truth solution $s$, and the sampled solution $\hat{s}$ from our LLM, the token rewritten probabilities from the generative reward model are used as the reward scores for the tokens in $\hat{s}$. 
To better indicate the token quality, we normalize the reward scores by subtracting them from the median value of the probability (\ie 0.5) and then clip the extreme values as:
\begin{equation}
\small
\label{reward_of_each_token}
R_{\hat{s}, t_j} = \textsc{Clip}(P_{RM}\big(t_{j}|p_{R},q,s,\hat{s},t_{<j})-0.5, \alpha, \beta\big),
\end{equation}
where $P_{RM}(t_{j}|p_{R},q,s,\hat{s},t_{<j})$ is the predicted probability of the correct token from the reward model for the $j$-th token in $\hat{s}$, and $\alpha$ and $\beta$ denote the minimum and maximum thresholds for the reward value.
For implementation, we employ $\alpha=-0.1$ and $\beta=0$ for the negative samples while adopt $\alpha=0$ and $\beta=0.5$ for the positive samples.
In this way, for negative samples, the upper threshold $\beta=0$ would lead to zero reward scores for all the non-error tokens, making the policy model only focus on punishing the error tokens. 
Otherwise, for positive samples, the lower threshold $\alpha=0$ would assign zero reward score to error tokens, enabling the policy to focus on learning the correct tokens.

\paratitle{Token-level RL Objective.}
Given the token-level reward scores, we perform RL training on the policy model to correct its behaviors to avoid making errors.
As mentioned in Section~\ref{preliminary}, we incorporate the PPO framework for RL, and revise its loss function to incorporate token-level reward scores.
Concretely, we aim to maximize the expectation that generates the desired correct tokens in the solution.
Thus, the gradients to optimize the policy model is given as:
\begin{equation}
\label{RLMEC_reward}
\small
\nabla \mathcal{J}_{{RL}}(\theta) = \sum_{i=1}^{n}\sum_{t_j \in \hat{s}_i} r(q_i,t_{j}) \times R_{\hat{s}_{i}, t_j} \times \nabla \log P_{\theta}(t_{j}|q_i,t_{<j}), \\
\end{equation}
where $\theta$ is the parameters of the policy model, $P_{\theta}(t_{j}|q_i,t_{<j})$ is the predicted probability of the $j$-th token by the policy model, and $r(q_i,t_{j})$ is the coefficient of the importance sampling in PPO as:
\begin{equation}
\small
r(q_i,t_{j}) = \frac{P_{\theta}(t_{j}|q_i,t_{<j})}{P_{\theta'}(t_{j}|q_i,t_{<j})}.
\end{equation}
Moreover, inspired by existing work~\cite{ppo,ppo_clip} that clips the gradients of RL, we design a simplified way that clips the coefficient of the gradient to reduce the variance of the reward and prevent the large difference between the policy and reference model:
\begin{equation}
\label{eq-clip}
\small
    \min\big(r(q_i,t_{j}) \times R_{\hat{s}_i,t_j}, \textsc{Clip}(r(q_i,t_{j}), 1-\varepsilon, 1+\varepsilon)\times R_{\hat{s}_i,j}\big),
\end{equation}
where $\varepsilon$ is a hyperparameter that controls the upper and lower bounds for positive and negative reward scores, respectively.

\begin{table}[t]
\small
    \centering
    \begin{tabular}{lcccc}
        \toprule
        \textbf{Methods} & \textbf{NS} & \textbf{RL} & \textbf{TLS} & \textbf{RM} \\
        \midrule
        SFT~\cite{instructgpt} & \ding{55} & \ding{55} & \ding{55} & - \\
        RFT~\cite{rft} & \ding{55} & \ding{55} & \ding{55} & DIS \\
        \midrule
        CoH~\cite{coh} & \ding{52} & \ding{55} & \ding{55} & - \\
        DPO~\cite{dpo} & \ding{52} & \ding{55} & \ding{55} & - \\
        FIGA~\cite{figa} & \ding{52} & \ding{55} & \ding{52} & DIS \\
        PPO~\cite{ppo} & \ding{52} & \ding{52} & \ding{55} & DIS \\
        \midrule
        ToRA~\cite{tora} & \ding{52} & \ding{55} & \ding{55} & - \\
        Shep.~\cite{math-shepherd} & \ding{52} & \ding{52} & \ding{55} & DIS \\
        WMath~\cite{wizardmath} & \ding{52} & \ding{52} & \ding{55} & DIS \\
        \midrule
        RLMEC & \ding{52} & \ding{52} & \ding{52} & GEN \\
        \bottomrule
    \end{tabular}
    \caption{The difference between RLMEC and previous related work. NS, RL, and TLS denote the usage of negative samples, reinforcement learning, and token-level supervision. RM denotes the type of the reward model. DIS and GEN denote the discriminative reward model and generative reward model, respectively.}
    \label{comparision}
\end{table}

\paratitle{Imitation-based Regularization.}
As the RL training process is prone to be unstable, we further design a regularization loss based on imitation learning. 
The policy model is trained to imitate the generation of the rewritten solution $\tilde{s}_i$, only based on the question $q_i$.
To compute the regularization term, we sample the generated wrong outputs $\hat{s}$ from the policy model, and utilize our generative reward model to rewrite it into a correct one $\Tilde{s}$ for learning. 
As discussed before, the original solution $\hat{s}_i$ may contain only few error tokens that lead to the wrong solution.
Therefore, we consider focusing on these error tokens in $\hat{s}$, and identify them for targeted learning.
Specifically, we leverage the Levenshtein Distance algorithm~\cite{Levenshtein-Distance}, an effective method to find the revised tokens in $\hat{s}$, and employ the token-level weights to emphasize them.
The Levenshtein Distance algorithm utilizes dynamic programming (DP) to calculate the edit distance between $\hat{s}$ and $\tilde{s}$, and the replaced and added tokens are selected into the error token set $\mathcal{T}$.
Then, the token-level weight is computed as:
\begin{equation}
\small
\label{eq-weight}
    w_{j} = \begin{cases} \gamma,&t_j \in \mathcal{T} \\ \phi \times \gamma,&t_j \notin \mathcal{T} \end{cases} ,
\end{equation}
where $\gamma$ denotes the weight for emphasized tokens in $\mathcal{T}$, and $\phi$ is the penalty coefficient for unimportant tokens. By incorporating term-level weights, the gradients of the imitation regularization are:
\begin{equation}
\small
\label{eq-imitation}
\nabla \mathcal{L}_{{IR}}(\theta) = -\sum_{i=1}^{n}\sum_{t_j \in \tilde{s}_i} \nabla \log P_{\theta}(t_{j}|q_i,t_{<j}) \times w_{j}.
\end{equation}
Finally, the policy model is optimized by both the RL objective and imitation-based regularization.

\ignore{
Typically, we define $D(x,y)$ to be the edit distance between the first $x$ tokens in $\hat{s}_i$ and the first $y$ tokens in $\tilde{s}_i$.
To guarantee the correctness of our algorithm, we present the proof of the optimal substructure in Appendix~\ref{}.
Therefore, the optimal substructure of the edit-distance problem gives the recursive formula,
\begin{equation}
\small
    D(x,y)=\text{min}\{D(x-1,y),D(x,y-1),D(x-1,y-1)\}+1
\end{equation}
where $D(x-1,y)$, $D(x,y-1)$, and $D(x-1,y-1)$ denote deletion, addition, and replacing, respectively.
Specially, in the situation of $\hat{s}_i[x]=\tilde{s}_i[y]$ (\ie the $x$-th token of $\hat{s}_i$ is the same as the $y$-th token of $\tilde{s}_i$), the transferring from $D(x-1,y-1)$ to $D(x,y)$ would not bring any expenses, \ie
\begin{equation}
\small
    D(x,y)=\text{min}\{D(x-1,y)+1,D(x,y-1)+1,D(x-1,y-1)\}
\end{equation}
During the DP process, we utilize $G(x,y)$ to record the state from which $D(x,y)$ has transferred.}

\subsection{Summary and Discussion}
Here, we present the summary of our approach and discuss its difference with existing methods.

\paratitle{Summary.}
We present the pseudo-code of RLMEC in Algorithm~\ref{code_rlmec} to better demonstrate our approach.
The procedure of RLMEC can be divided into two parts, \ie generative reward model training and reinforcement learning with fine-grained supervision.
For generative reward model training, we leverage a teacher model (\ie Claude 2) to synthesize the examples for the error locating and solution rewriting subtasks, to compose the dataset for distilling our generative reward model the capability of erroneous solution rewriting.
Then, for RL training, we first generate the rewards for all the tokens in the sampled solutions from the policy model using Eq.~\ref{reward_of_each_token}, where we set suitable thresholds $\alpha$ and $\beta$ to control our model to focus on important tokens in the generated solutions.
Based on the token-level reward, we perform RL training using the PPO framework with the optimization function Eq.~\eqref{RLMEC_reward}, and we design the reward clip strategy using Eq.~\eqref{eq-clip} to prevent extreme rewards and stabilize the training process.
Besides, we also add the imitation-based regularization using Eq.~\eqref{eq-imitation}, to further help our policy model focus on learning key tokens.


\paratitle{Discussion.}
In Tabel~\ref{comparision}, we present the difference between RLMEC and the existing work.
Previous work mostly adopts the instance-level reward model, and only FIGA employs the token-level supervision but does not utilize RL. 
Besides, there are several methods (\eg WizardMath, Math-Shepherd) that leverage step-level reward to perform RL.
As a comparison, our proposed RLMEC enables token-level supervision in the RL framework, and thus can benefit from more fine-grained supervision and focus on punishing error tokens during training procedure. 
A major novelty of our implementation is that we design the generative reward model trained by the erroneous solution rewriting task, to replace the conventional discriminative reward model, which can produce a rewritten probability of each token that can be naturally used as token-level supervision. 
Besides, by comparing with supervised fine-tuning methods (\eg SFT and RFT), our approach can utilize the negative samples that will not be used by them, which extends the understanding of failed examples and fully utilizes the data.
\section{Experiment}


\subsection{Experimental Settings}

\begin{table}[t]
\small
    \centering
    \begin{tabular}{cccc}
        \toprule
        \textbf{Task} & \textbf{Train/Test} & \textbf{Dataset} & \textbf{Num. Data} \\
        \midrule
        \multirow{5.5}*{Math} & Train & MathInst & 118088 \\
            \cmidrule{2-4}
            & \multirow{4}*{Test} & GSM8k & 1319 \\
            &  & MATH & 5000 \\
            &  & SVAMP & 1000 \\
            &  & MM & 974 \\
        \midrule
        \multirow{6.5}*{QA} & \multirow{2}*{Train} & ECQA & 7598 \\
                & & QASC & 8134 \\
            \cmidrule{2-4}
            & \multirow{4}*{Test} & ECQA & 2194 \\
                & & QASC & 926 \\
                & & OBQA & 500 \\
                & & ARC & 2376 \\
        \bottomrule
    \end{tabular}
    \caption{Statistics of the used datasets. MathInst and MM denote MathInstruct and the mathematical task in MMLU, respectively.}
    \label{dataset}
\end{table}

We simply introduce the experimental settings in this part. More details are shown in Appendix~\ref{implementation_details}.

\paratitle{Datasets.}
We employ mathematical tasks and question-answering tasks for evaluation. 
The specifics of each dataset are delineated in Table~\ref{dataset}.
Mathematical tasks include GSM8k~\cite{gsm8k}, MATH~\cite{math}, SVAMP~\cite{svamp} and the mathematical problems in MMLU (MM)~\cite{mmlu_1,mmlu_2}. 
We adopt MathInstruct~\cite{mammoth} as the training set and eliminate the code samples. 
Question-answering tasks contain ECQA~\cite{ecqa}, QASC~\cite{qasc}, OpenbookQA~\cite{openbookqa} and ARC-Easy~\cite{arc}.
We merge the training set of ECQA and QASC, and adopt the mixture as the training set in the experiment.

\begin{table*}[ht]
    \small
    \centering
    \begin{tabular}{lcccccccccc}
        \toprule
        \multirow{2.5}*{\textbf{Methods}} & \multicolumn{5}{c}{\textbf{Question-Answering Tasks}} & \multicolumn{5}{c}{\textbf{Mathematical Tasks}} \\
        \cmidrule(r){2-6}\cmidrule(r){7-11}
         & ECQA & QASC & OBQA & ARC & Avg. & GSM8k & MATH & SVAMP & MM & Avg.  \\
        \midrule
        \multicolumn{11}{c}{\textit{7B Parameters LLMs}} \\
        LLaMA 2 & 55.97 & 39.74 & 48.40 & 52.48 & 49.15 & 11.22 & 4.80 & 29.70 & 28.44 & 18.54 \\
        Vicuna & 49.82 & 32.18 & 46.40 & 51.52 & 44.98 & 12.20 & 4.26 & 24.30 & 26.08 & 16.71 \\
        WizardLM & 36.28 & 18.68 & 27.80 & 46.59 & 32.34 & 14.48 & 3.34 & 34.80 & 27.10 & 19.93 \\
        SFT LLM & 71.88 & 55.40 & 52.00 & 56.27 & 58.89 & 51.02 & 10.48 & 47.80 & 38.50 & 36.95 \\
        + SFT & 70.65 & 55.94 & 51.60 & 56.99 & 58.80 & 50.34 & 11.04 & 47.20 & 38.40 & 36.75 \\
        + RFT & 72.24 & 58.64 & 55.20 & 57.15 & 60.81 & 49.66 & 10.80 & 48.30 & 39.01 & 36.94 \\
        + RFT w/ GT & 72.47 & 58.53 & 53.60 & 57.11 & 60.43 & 49.89 & \textbf{11.26} & 46.70 & 38.91 & 36.69 \\
        + RFT w/ TD & \underline{73.11} & 58.21 & 54.20 & \underline{57.53} & 60.76 & \textbf{51.86} & 11.04 & \underline{49.40} & 38.19 & 37.62 \\
        + RFT w/ RD & 72.47 & \underline{59.29} & 54.60 & 57.03 & \underline{60.85} & \underline{51.78} & \underline{11.24} & 48.70 & \underline{40.76} & \underline{38.12} \\
        + CoH & 71.06 & 54.86 & 51.40 & 56.61 & 58.48 & 50.11 & 10.94 & 48.60 & 38.50 & 37.04 \\
        + DPO & 72.47 & 58.53 & \underline{55.40} & 55.26 & 60.42 & 34.19 & 5.38 & 25.80 & 32.58 & 24.49 \\
        + FIGA & 69.83 & 52.48 & 51.00 & 46.21 & 54.88 & - & - & - & - & - \\
        + Vanilla PPO & 72.88 & 50.22 & 43.40 & 56.27 & 55.69 & 48.97 & 10.64 & 44.90 & 38.60 & 35.78 \\
        + PPO A2C & 70.83 & 55.08 & 52.40 & 56.02 & 58.58 & 50.94 & 9.38 & 46.60 & 38.50 & 36.36 \\
        + RLMEC & \textbf{73.66} & \textbf{59.50} & \textbf{56.80} & \textbf{58.50} & \textbf{62.12} & 51.18 & 11.16 & \textbf{49.60} & \textbf{40.97} & \textbf{38.23} \\
        \midrule
        \multicolumn{11}{c}{\textit{13B Parameters LLMs}} \\
        LLaMA 2 & 61.53 & 45.46 & 57.90 & \underline{64.31} & 57.30 & 21.23 & 6.58 & 34.40 & 34.39 & 24.15 \\
        Vicuna & 50.14 & 39.96 & 48.40 & 53.70 & 48.05 & 24.10 & 4.74 & 33.80 & 29.98 & 23.16 \\
        WizardLM & 52.60 & 40.93 & 52.30 & 58.96 & 51.20 & 31.01 & 3.18 & 52.00 & 21.36 & 26.89 \\
        SFT LLM & 76.12 & 59.40 & 60.80 & 62.46 & 64.70 & 56.63 & 12.74 & 53.50 & 41.27 & 41.04 \\
        + SFT & 75.89 & 57.87 & \underline{63.40} & 62.50 & 64.92 & 55.88 & 13.62 & 58.00 & 41.27 & 42.19 \\
        + RFT & 75.71 & 60.48 & 61.00 & 64.06 & 65.31 & 55.80 & 13.62 & 54.10 & 41.68 & 41.30 \\
        + RFT w/ GT & 76.66 & 60.37 & \underline{63.40} & 63.17 & 65.90 & 57.32 & 13.74 & 56.70 & 43.94 & 42.93 \\
        + RFT w/ TD & 76.71 & 61.56 & 61.80 & 64.14 & 66.05 & \textbf{58.15} & 13.98 & \underline{58.80} & 41.58 & \underline{43.13} \\
        + RFT w/ RD & 76.62 & \underline{62.20} & 63.20 & 63.17 & 66.30 & 57.39 & \textbf{14.34} & 56.20 & \underline{42.81} & 42.96 \\
        + CoH & 76.62 & 60.37 & 59.80 & 63.93 & 65.18 & 57.31 & 13.10 & 54.00 & 42.30 & 41.68 \\
        + DPO & \underline{78.26} & 61.45 & 62.20 & 63.80 & \underline{66.43} & 44.20 & 4.38 & 39.70 & 32.14 & 30.11 \\
        + FIGA & 61.21 & 60.26 & 52.80 & 46.34 & 55.15 & - & - & - & - & - \\
        + Vanilla PPO & 76.34 & 57.99 & 61.80 & 62.29 & 64.61 & 53.45 & 11.76 & 55.10 & 43.12 & 40.86 \\
        + RLMEC & \textbf{79.49} & \textbf{64.15} & \textbf{65.60} & \textbf{65.19} & \textbf{68.61} & \textbf{58.15} & \underline{14.00} & \textbf{60.00} & \textbf{45.07} & \textbf{44.31} \\
        \bottomrule
    \end{tabular}
    \caption{Experimental results on question answering tasks and mathematical tasks. Avg. is the average accuracy of all sub-tasks. GT, TD, and RD denote ground truth, the data generated by the teacher model, and the data generated by the generative reward model. The best are denoted in bold and the second-best are underlined.}
    \label{main_results}
\end{table*}

\paratitle{Baselines.}
For a more comprehensive assessment, we incorporate three categories of methods as baseline approaches.
We conduct the SFT~\cite{instructgpt} and the Rejection sampling Fine-Tuning~(RFT)~\cite{rejection_sampling,rft} as the baseline methods of supervised fine-tuning.
To conduct the more persuasive experiment, we also evaluate the variants of RFT, including adding the ground truth solution, rewritten solution from the teacher model, and rewritten solution from the generative reward model, named RFT w/ GT, RFT w/ TD, and RFT w/ RD, respectively.
Besides, the representative methods of alignment without reinforcement learning, \eg DPO~\cite{dpo}, CoH~\cite{coh}, and FIGA~\cite{figa} are conducted as the baseline.
Moreover, We conduct the vanilla PPO~\cite{ppo} and Actor-Critic version of PPO (PPO A2C)~\cite{moss_rlhf} as the baseline of RL methods.
Additionally, we also report the performance of base LLMs, including LLaMA 2~\cite{llama2}, Vicuna~\cite{vicuna}, and WizardLM~\cite{wizardlm}.

\subsection{Main Results}

The evaluation results of RLMEC and the baseline methods are presented in Table~\ref{main_results}.

First, RLMEC outperforms other baselines on the average accuracy of both scenarios.
RLMEC demonstrates a strong capacity to further enhance the specific ability (\eg reasoning ability) of LLMs. 
With the limited training data, compared with the previous methods (\eg RFT, PPO), RLMEC leverages both positive and negative samples to provide fine-grained supervision signals, guiding LLMs to focus on the mistakes and correct them.

Second, RLMEC can prevent overfitting during domain adaption.
Previous methods (\eg SFT) utilize the data from the training set or generated by LLMs to fine-tune the LLMs which might cause overfitting.
We can observe that the performance decreases after SFT on the unseen tasks (\eg OBQA and SVAMP) of the 7B LLM.
In contrast, the performance of LLMs on all of the unseen tasks is improved after RLMEC.
The reason is that RLMEC makes LLMs focus on mistakes rather than correct components and utilize the clip mechanism to avoid overfitting.

Third, RLMEC can better leverage the generated response containing undesired components than other methods.
Comparing the performance of RLMEC and DPO, we can observe that RLMEC enhance the reasoning ability of LLMs in both scenarios, but DPO only works on question-answer tasks.
That is because RLMEC utilizes soft rewards to indicate positive or negative responses, while DPO collects the positive-negative response pairs to train LLMs which can be regarded as utilizing the hard labels to identify the quality of generated responses.
Given the quality of generated responses is difficult to assess, it is hard to collect response pairs in the challenge tasks (\eg mathematical tasks).
On mathematical tasks, the performance of DPO is even worse than the backbone LLM because of the low quality of the training data.

Finally, token-level supervision signals can further improve the performance of the policy model.
The results of vanilla PPO, PPO A2C, and RLMEC present the importance of fine-grained supervision signals.
Vanilla PPO utilizes instance-level signals to train the LLMs, which do not conform to reality because the generated response might contain both desired and undesired components.
PPO A2C trains the critic model to provide fine-grained supervision signals which will increase the requirement of the computation resources.
In RLMEC, the generative reward model is competent to implement the functionality of the reward model and the critic model in the PPO A2C at the same time.

\subsection{Detailed Analysis}

\begin{table}[t]
\small
    \centering
    \begin{tabular}{ccccccc}
        \toprule
        \multicolumn{3}{c}{\textbf{Methods}} & \textbf{ECQA} & \textbf{ARC} & \textbf{GSM8k} & \textbf{MM}  \\
        \cmidrule(r){1-3}\cmidrule(r){4-4}\cmidrule(r){5-5}\cmidrule(r){6-6}\cmidrule(r){7-7}
        TLS & RL & IR & Acc. & Acc. & Acc. & Acc. \\
        \midrule
        \ding{52} & \ding{52} & \ding{52} & 79.49 & 65.19 & 58.15 & 45.07 \\
        \ding{55} & \ding{52} & \ding{52} & 78.81 & 64.52 & 58.38 & 44.45 \\
        \ding{55} & \ding{55} & \ding{52} & 77.85 & 64.18 & 58.56 & 43.84 \\
        \ding{52} & \ding{52} & \ding{55} & 74.34 & 61.32 & 7.35 & 20.12 \\
        \bottomrule
    \end{tabular}
    \caption{The results of ablation study on 13B LLMs. TLS, RL, and IR denote token-level supervision, reinforcement learning, and imitation-based regularization.}
    \label{ablation}
\end{table}




To further verify the effectiveness of RLMEC, we conduct the ablation study and analyze the model performance during the training process. Besides, we analyze the scaling of the generative reward model and present the case study of supervision signals and the model outputs in Appendix~\ref{analysis} and~\ref{case_study}.

\paratitle{Ablation Study.}
We evaluate the effectiveness of token-level supervision, reinforcement learning, and imitation-based regularization. Results are presented in Table~\ref{ablation}.
Given the results of the QA tasks (\ie ECQA and ARC), we can observe that removing any of the modules will hurt the performance of the LLMs.
In the mathematical tasks, without token-level supervision and reinforcement learning, LLMs overfit the training set, which brings the improvement on the seen task (\ie GSM8k) and hurts the performance on the unseen task (\ie MM).
The evaluation results demonstrate the ability of RLMEC to prevent overfitting and achieve the balance between seen tasks and unseen tasks.
Besides, imitation-based regularization is also an important module in RLMEC.
Without regularization, LLMs learn to generate correct responses only through token-level rewards.
Because of the large search space, it is very difficult for LLMs to find the correct behavior in the challenge tasks.
In the setting of removing imitation-based regularization, the decreasing performance on all of the tasks can verify our analysis.


\begin{figure}[t]
    \centering
    \includegraphics[width=0.48\textwidth]{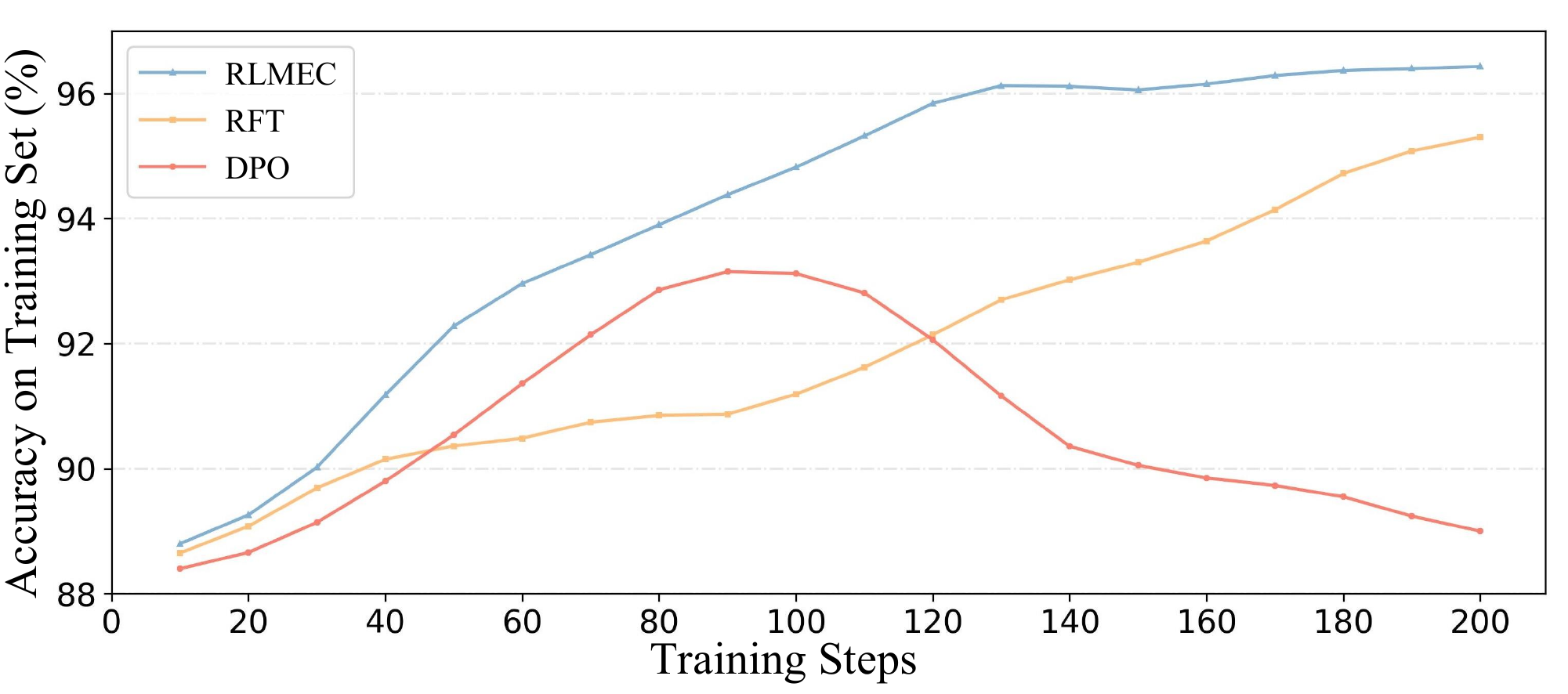}
    \caption{The performance of 7B LLMs on question-answering tasks during different training strategies. To better present the difference, we smooth out the lines.}
    \label{train_acc}
\end{figure}

\paratitle{Performance During Training Process.}
To comprehensively assess the performance of RLMEC, we conduct experiments on the accuracy of the training set during the training process.
In Figure~\ref{train_acc}, we can observe that RLMEC can fit the training set more effectively and rapidly than other methods (\ie RFT and DPO).
Around 120 training steps, the policy model almost fits the training set through RLMEC.
That is because our methods focus on the mistakes in the generated response and guide LLMs to correct these errors, which is more efficient.
In contrast, RFT optimizes the whole tokens in the correct solution which might include many unimportant tokens, and DPO is overemphasized about the negative samples.
These futures will decrease the speed of optimization and hurt the performance.


\section{Conclusion}
In this paper, we proposed RLMEC, a new reinforcement learning framework with minimum editing constraint, to leverage fine-grained supervision signals to further improve the ability of LLMs.
In our RLMEC, we first trained the generative reward model via the erroneous solution rewriting task under the minimum editing constraint, with the help of a teacher LLM. 
Then, we leveraged it to produce token-level rewards, and devised the token-level RL objective and an imitation-based regularization for training our LLM, which both focus on the revision of the key tokens leading to errors in the solution.
Experimental results on mathematical tasks and question-answering tasks have demonstrated the effectiveness of RLMEC.

As future work, we will consider implementing our RL method on more advanced LLMs to further improve their performance on complex reasoning tasks. Besides, we will also evaluate the capacity of our approach on enhancing human alignment and reducing hallucination.

\section*{Limitations}
In this section, we discuss the limitations of our work.
First, in this work, we focus on the complex reasoning tasks and only conduct experiments on the QA tasks and mathematical tasks.
However, RLMEC can also be employed in other scenarios, \eg human alignment and reducing hallucination, which has not been verified in this work.
We leave it as the future work.
Second, due to the limitation of computing resources, we only assess the performance of RLMEC on 7B and 13B LLMs, without the experiments on larger LLMs.
Actually, by comparing the performance of baseline methods and RLMEC on 7B and 13B LLMs, we can observe the effectiveness of RLMEC.
Third, our approach mainly focuses on enhancing LLMs on complex reasoning tasks, and does not consider the possible bias and ethic risks when using LLMs.
It is also a promising direction that our RLMEC can be applied to, and we will investigate it in the future.

\section*{Acknowledgement}
This work was partially supported by National Natural Science Foundation of China under Grant No. 62222215, Beijing Natural Science Foundation under Grant No. L233008 and 4222027. Xin Zhao is the corresponding author.


\bibliography{anthology,custom}
\bibliographystyle{acl_natbib}

\newpage

\appendix

\begin{table*}[t]
    \small
    \centering
    \begin{tabular}{ll}
        \toprule
        \multicolumn{1}{l}{\begin{tabularx}{0.1\textwidth}{@{}X@{}}
            \textbf{Error}\\ \textbf{Locating}
        \end{tabularx}} & \multicolumn{1}{l}{\begin{tabularx}{0.8\textwidth}{@{}X@{}}
            Given the problem, correct solution and the prediction from language models. The method in prediction might be different with correct solution, but it is also correct. You need to identify which step of the prediction is the first wrong step, and write down the label of the first wrong step. \\ \\Problem: \{\texttt{Problem}~$q$\} \\ \\ Correct solution: \{\texttt{Formatted Ground-Truth Solution}~$s$\} \\ \\ Prediction: \{\texttt{Generated Erroneous Solution}~$\hat{s}$\} \\ \\ Which step of prediction is error? Only write down the label of the first wrong step. If the prediction is correct, you need to write down correct. You should not write down any other words.
        \end{tabularx}} \\
        \midrule
        \multicolumn{1}{l}{\begin{tabularx}{0.1\textwidth}{@{}X@{}}
            \textbf{Solution Rewriting}
        \end{tabularx}} & \multicolumn{1}{l}{\begin{tabularx}{0.8\textwidth}{@{}X@{}}
            Given the problem and the correct solution, you need to correct the mistakes in prediction to get the correct answer. You should make minimal modifications. \\ \\ Problem: \{\texttt{Problem}~$q$\} \\ \\ Correct solution: \{\texttt{Generated Erroneous Solution}~$\hat{s}$\} \\ \\ Prediction: \{\texttt{Generated Erroneous Solution}~$\hat{s}$\} \\ \\ Correct prediction: 
        \end{tabularx}} \\
        \bottomrule
    \end{tabular}
    \caption{The prompt for the teacher model distillation.}
    \label{prompt_tm}
\end{table*}

\begin{table*}[t]
    \small
    \centering
    \begin{tabular}{ll}
        \toprule
        \multicolumn{1}{l}{\begin{tabularx}{0.1\textwidth}{@{}X@{}}
            \textbf{Error}\\ \textbf{Locating}
        \end{tabularx}} & \multicolumn{1}{l}{\begin{tabularx}{0.8\textwidth}{@{}X@{}}
            Below is an instruction that describes a task. Write a response that appropriately completes the request.\\ \#\#\# Instruction:\\ Given the problem, correct solution and the prediction from language models. The method in prediction might be different with correct solution, but it is also correct. You need to identify which step of the prediction is the first wrong step, and write down the label of the first wrong step.\\ \\ \#\#\# Input:\\ Problem: \{\texttt{Question}~$q$\} \\Correct solution: \{\texttt{Formatted Ground-Truth Solution}~$s$\}\\Prediction: \{\texttt{Generated Erroneous Solution}~$\hat{s}$\}\\ \\ \#\#\# Response:\\ \textbf{The first error step is [\{\texttt{First Undesired Reasoning Step}~$r_t$\}]}
        \end{tabularx}} \\
        \midrule
        \multicolumn{1}{l}{\begin{tabularx}{0.1\textwidth}{@{}X@{}}
            \textbf{Solution Rewriting}
        \end{tabularx}} & \multicolumn{1}{l}{\begin{tabularx}{0.8\textwidth}{@{}X@{}}
            Below is an instruction that describes a task. Write a response that appropriately completes the request.\\ \#\#\# Instruction:\\ Given the problem and the correct solution, you need to correct the mistakes in prediction to get the correct answer. You should make minimal modifications.\\ \\ \#\#\# Input:\\ Problem: \{\texttt{Question}~$q$\} \\Correct solution: \{\texttt{Ground-Truth Solution}~$s$\}\\Prediction: \{\texttt{Generated Erroneous Solution}~$\hat{s}$\}\\ \\ \#\#\# Response:\\ Correct prediction:\textbf{\{\texttt{refined solution}~$\tilde{s}$\}}
        \end{tabularx}} \\
        \bottomrule
    \end{tabular}
    \caption{The instruction for the generative reward model training. The bold sentence will be utilized to optimize the generative reward model in cross entropy loss. The prompt for inference is the same as the training instruction without the bold part.}
    \label{prompt_rm}
\end{table*}

\begin{figure}[h]
	\centering
	\begin{subfigure}{\linewidth}
	    \centering
		\includegraphics[width=\linewidth]{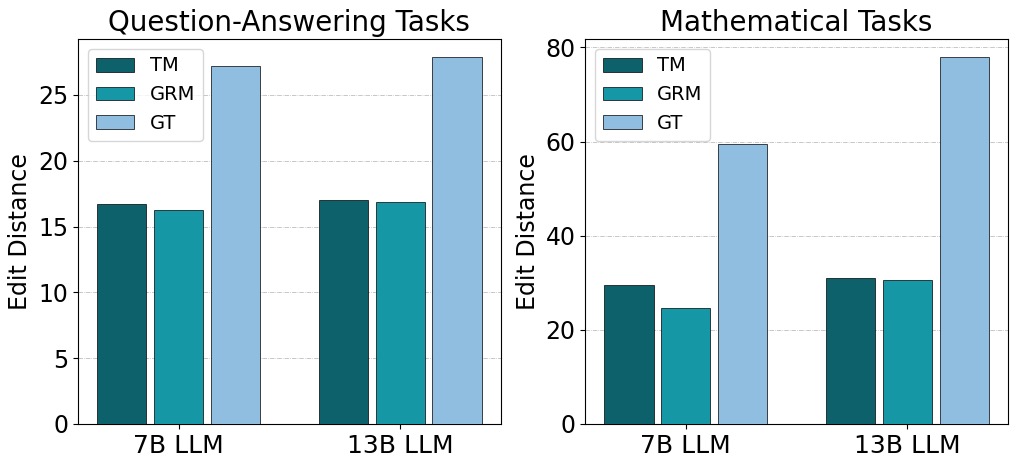}
		\caption{Edit distance between refined responses and predictions.}
		\label{edit_dist}
	\end{subfigure}

	\begin{subfigure}{\linewidth}
            \centering
		\includegraphics[width=\linewidth]{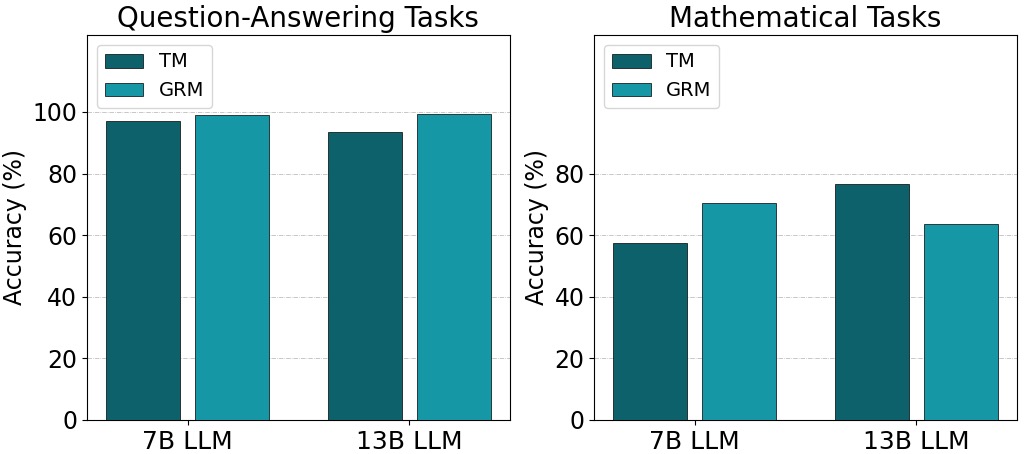}
		\caption{Accuracy of the refined response.}
		\label{fix_acc}
	\end{subfigure}
        \caption{The comparison of the rewriting performance of teacher model and generative reward model. TM and GRM denote the response refined by the teacher model and the generative reward model, respectively. GT denotes the ground truth solution of the problems.}
        \label{tm_rm}
\end{figure}

\begin{figure}[h]
    \centering
    \includegraphics[width=0.48\textwidth]{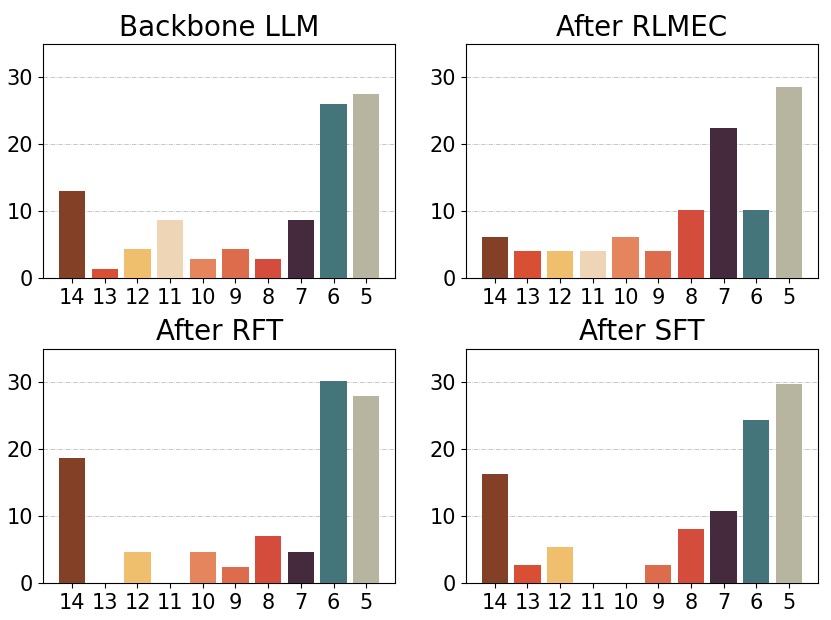}
    \caption{The position of the first error in the generated solution. The X-axis denotes how many reasoning steps between the first error and the final answer, and the Y-axis is the ratio of the corresponding problems in these problems.}
    \label{error_shift}
\end{figure}

\begin{table}[t]
\small
    \centering
    \begin{tabular}{ccccc}
        \toprule
        \multirow{2.5}*{\diagbox{\textbf{GRM}}{\textbf{PM}}} &  \multicolumn{2}{c}{\textbf{7B PM}} & \multicolumn{2}{c}{\textbf{13B PM}}  \\
        \cmidrule(r){2-3}\cmidrule(r){4-5}
        & QA & Math & QA & Math \\
        \midrule
        \textbf{7B GRM} & 62.12 & 38.23 & 66.40 & 43.74 \\
        \textbf{13B GRM} & 61.32 & 37.46 & 68.61 & 44.31 \\
        \bottomrule
    \end{tabular}
    \caption{The comparison of the different scaling of the generative reward model. GRM and PM denote the generative reward model and policy model, respectively.}
    \label{scale_of_rm}
\end{table}

\begin{algorithm}[t]
\small
\caption{The RLMEC algorithm.}
\label{code_rlmec}
\SetKwInOut{Input}{Input}
\SetKwInOut{Output}{Output}

\Input{Training set $\mathcal{D}=\{\langle q_i,s_i \rangle\}_{i=1}^{n}$, the teacher model (Claude 2), and the parameters of SFT model $\theta_{SFT}$.}
\Output{A well trained policy model.}

\BlankLine
Initialize the parameters of the generative reward model: $\theta_{GRM} \leftarrow \theta_{SFT}$\;
Initialize the parameters of the policy model: $\theta \leftarrow \theta_{SFT}$\;

\BlankLine
\tcp{Generative Reward Model Training}
\For{each instance $\langle q_i,s_i \rangle$ in $\mathcal{D}$}{
    The policy model generates $\hat{s}_i$ based on $q_i$\;
    \If{the data is sampled}{
        The teacher model locals the first error at $r_i$ using Eq.~\ref{eq-locate}\;
        The teacher model rewrites $\hat{s}_i$ to obtain $\tilde{s}_i$ using Eq.~\ref{eq-rewrite}\;
    }
}
Use $q_i$, $s_i$, $\hat{s}_i$, $r_i$, and $\tilde{s}_i$ to construct $\mathcal{D}'$\;
Leverage $\mathcal{D}'$ to supervised-finetune the generative reward model through Seq2Seq training paradigm\;

\BlankLine
\tcp{RL with Fine-grained Supervision}
\For{each instance $\langle q_i,s_i,\hat{s}_i,r_i,\tilde{s}_i \rangle$ in $\mathcal{D}'$}{
    Generate the token-level rewards using Eq.~\ref{reward_of_each_token}\;
    Compute the reward $\mathcal{J}_{RL}(\theta)$ using Eq.~\ref{RLMEC_reward}\;
    Use Levenshtein Distance algorithm to compute the token-level weight\;
    Compute the loss of imitation-based regularization $\mathcal{L}_{IR}(\theta)$\;
    Update $\theta$ through $\mathcal{J}_{RL}(\theta)$ and $\mathcal{L}_{IR}(\theta)$\;
}
\end{algorithm}

\definecolor{e1}{RGB}{255, 154, 154}
\definecolor{e2}{RGB}{255, 174, 174}
\definecolor{e3}{RGB}{255, 194, 194}
\definecolor{e4}{RGB}{255, 214, 214}
\definecolor{e8}{RGB}{204, 255, 204}

\begin{table*}[ht]
    \small
    \centering
    \begin{tabular}{ll}
        \toprule
        \multicolumn{1}{l}{\begin{tabularx}{0.1\textwidth}{@{}X@{}}
            \textbf{Problem}
        \end{tabularx}} & \multicolumn{1}{l}{\begin{tabularx}{0.8\textwidth}{@{}X@{}}
            Look at this series: 1.25, 7.5, \_\_\_\_, 270, 1620, ... What number should fill the blank? \\ Answer Choices: (A) 50 (B) 65 (C) 45 (D) 55 (E) 40
        \end{tabularx}} \\
        \midrule
        \multicolumn{1}{l}{\begin{tabularx}{0.12\textwidth}{@{}X@{}}
            \textbf{Ground Truth Solution}
        \end{tabularx}} & \multicolumn{1}{l}{\begin{tabularx}{0.8\textwidth}{@{}X@{}}
            Let's think about the multi-choice question. \\ This is a simple multiplication series. Each number is 6 times greater than the previous number. \\ The answer is C
        \end{tabularx}} \\
        \midrule
        \multicolumn{1}{l}{\begin{tabularx}{0.12\textwidth}{@{}X@{}}
            \textbf{Generated Solution}
        \end{tabularx}} & \multicolumn{1}{l}{\begin{tabularx}{0.8\textwidth}{@{}X@{}}
            Let's think about the multi-choice question step by step. \\ This is a simple multiplication series. Each number is 6 times greater than the previous number. \\ The answer is \textcolor{red}{B}
        \end{tabularx}} \\
        \midrule
        \multicolumn{1}{l}{\begin{tabularx}{0.12\textwidth}{@{}X@{}}
            \textbf{Rewards from RLMEC}
        \end{tabularx}} & \multicolumn{1}{l}{\begin{tabularx}{0.8\textwidth}{@{}X@{}}
            \sethlcolor{e8}\hl{Let}\sethlcolor{e8}\hl{'}\sethlcolor{e8}\hl{s~}\sethlcolor{e8}\hl{think~}\sethlcolor{e8}\hl{about~}\sethlcolor{e8}\hl{the~}\sethlcolor{e8}\hl{multi}\sethlcolor{e8}\hl{-}\sethlcolor{e8}\hl{choice~}\sethlcolor{e8}\hl{question~}\sethlcolor{e8}\hl{step~}\sethlcolor{e8}\hl{by~}\sethlcolor{e8}\hl{step~}\sethlcolor{e8}\hl{.} \\ \sethlcolor{e8}\hl{This~}\sethlcolor{e8}\hl{is~}\sethlcolor{e8}\hl{a~}\sethlcolor{e8}\hl{simple~}\sethlcolor{e8}\hl{multiplication~}\sethlcolor{e8}\hl{series}\sethlcolor{e8}\hl{.~}\sethlcolor{e8}\hl{Each~}\sethlcolor{e8}\hl{number~}\sethlcolor{e8}\hl{is~}\sethlcolor{e8}\hl{~}\sethlcolor{e8}\hl{6~}\sethlcolor{e8}\hl{times~}\sethlcolor{e8}\hl{greater~}\sethlcolor{e8}\hl{than~}\sethlcolor{e8}\hl{the~}\sethlcolor{e8}\hl{previous~}\sethlcolor{e8}\hl{number~}\sethlcolor{e8}\hl{.} \\ \sethlcolor{e8}\hl{The~}\sethlcolor{e8}\hl{answer~}\sethlcolor{e8}\hl{is~}\sethlcolor{e1}\hl{B}
        \end{tabularx}} \\
        \midrule
        \multicolumn{1}{l}{\begin{tabularx}{0.12\textwidth}{@{}X@{}}
            \textbf{Rewards from PPO A2C}
        \end{tabularx}} & \multicolumn{1}{l}{\begin{tabularx}{0.8\textwidth}{@{}X@{}}
            \sethlcolor{e1}\hl{Let}\sethlcolor{e1}\hl{'}\sethlcolor{e1}\hl{s~}\sethlcolor{e1}\hl{think~}\sethlcolor{e1}\hl{about~}\sethlcolor{e1}\hl{the~}\sethlcolor{e1}\hl{multi}\sethlcolor{e1}\hl{-}\sethlcolor{e1}\hl{choice~}\sethlcolor{e1}\hl{question~}\sethlcolor{e1}\hl{step~}\sethlcolor{e1}\hl{by~}\sethlcolor{e1}\hl{step~}\sethlcolor{e1}\hl{.} \\ \sethlcolor{e1}\hl{This~}\sethlcolor{e1}\hl{is~}\sethlcolor{e1}\hl{a~}\sethlcolor{e1}\hl{simple~}\sethlcolor{e1}\hl{multiplication~}\sethlcolor{e1}\hl{series~}\sethlcolor{e2}\hl{.~}\sethlcolor{e2}\hl{Each~}\sethlcolor{e2}\hl{number~}\sethlcolor{e2}\hl{is~}\sethlcolor{e2}\hl{~}\sethlcolor{e2}\hl{6~}\sethlcolor{e2}\hl{times~}\sethlcolor{e3}\hl{greater~}\sethlcolor{e3}\hl{than~}\sethlcolor{e3}\hl{the~}\sethlcolor{e3}\hl{previous~}\sethlcolor{e3}\hl{number~}\sethlcolor{e3}\hl{.} \\ \sethlcolor{e4}\hl{The~}\sethlcolor{e4}\hl{answer~}\sethlcolor{e4}\hl{is~}\sethlcolor{e4}\hl{B~}
        \end{tabularx}} \\
        \midrule
        \multicolumn{1}{l}{\begin{tabularx}{0.12\textwidth}{@{}X@{}}
            \textbf{Rewards from Vanilla PPO}
        \end{tabularx}} & \multicolumn{1}{l}{\begin{tabularx}{0.8\textwidth}{@{}X@{}}
            \sethlcolor{e1}\hl{Let}\sethlcolor{e1}\hl{'}\sethlcolor{e1}\hl{s~}\sethlcolor{e1}\hl{think~}\sethlcolor{e1}\hl{about~}\sethlcolor{e1}\hl{the~}\sethlcolor{e1}\hl{multi}\sethlcolor{e1}\hl{-}\sethlcolor{e1}\hl{choice~}\sethlcolor{e1}\hl{qu}\sethlcolor{e1}\hl{estion~}\sethlcolor{e1}\hl{step~}\sethlcolor{e1}\hl{by~}\sethlcolor{e1}\hl{step~}\sethlcolor{e1}\hl{.} \\ \sethlcolor{e1}\hl{This~}\sethlcolor{e1}\hl{is~}\sethlcolor{e1}\hl{a~}\sethlcolor{e1}\hl{simple~}\sethlcolor{e1}\hl{multiplication~}\sethlcolor{e1}\hl{series~}\sethlcolor{e1}\hl{.}\sethlcolor{e1}\hl{Each~}\sethlcolor{e1}\hl{number~}\sethlcolor{e1}\hl{is~}\sethlcolor{e1}\hl{~}\sethlcolor{e1}\hl{6~}\sethlcolor{e1}\hl{times~}\sethlcolor{e1}\hl{greater~}\sethlcolor{e1}\hl{than~}\sethlcolor{e1}\hl{the~}\sethlcolor{e1}\hl{previous~}\sethlcolor{e1}\hl{number~}\sethlcolor{e1}\hl{.} \\ \sethlcolor{e1}\hl{The~}\sethlcolor{e1}\hl{answer~}\sethlcolor{e1}\hl{is~}\sethlcolor{e1}\hl{B~}
        \end{tabularx}} \\
        \bottomrule
    \end{tabular}
    \caption{The comparison of the reward of the generated solution from different methods. We use different background colors to indicate the reward. The color changing from red to green denotes the reward changing from \colorbox{e1}{negative} to \colorbox{e8}{positive}.}
    \label{reward}
\end{table*}

\begin{table*}[ht]
    \small
    \centering
    \begin{tabular}{ll}
        \toprule
        \multicolumn{1}{l}{\begin{tabularx}{0.1\textwidth}{@{}X@{}}
            \textbf{Problem}
        \end{tabularx}} & \multicolumn{1}{l}{\begin{tabularx}{0.8\textwidth}{@{}X@{}}
            Below is an instruction that describes a task. Write a response that appropriately completes the request.\\ \\ \#\#\# Instruction:\\Applying for a job can make someone feel what sort of emotion, \underline{even if they get it}?\\Options:\\working hard\\frustration\\rejection\\defeat\\stress \\ \\ \#\#\# Response: Let's think step by step.
        \end{tabularx}} \\
        \midrule
        \multicolumn{1}{l}{\begin{tabularx}{0.1\textwidth}{@{}X@{}}
            \textbf{Ground-Truth Solution}
        \end{tabularx}} & \multicolumn{1}{l}{\begin{tabularx}{0.8\textwidth}{@{}X@{}}
            Stress is a feeling of tension about uncertainty.\\Applying for a job has an uncertain outcome.\\The answer is stress
        \end{tabularx}} \\
        \midrule
        \multicolumn{1}{l}{\begin{tabularx}{0.1\textwidth}{@{}X@{}}
            \textbf{Generated Solution Before Training}
        \end{tabularx}} & \multicolumn{1}{l}{\begin{tabularx}{0.8\textwidth}{@{}X@{}}
            Applying for a job is followed by attending an interview.\\Attending an interview gives \textcolor{red}{rejection or acceptance}.\\Rejection is an emotion.\\The answer is rejection
        \end{tabularx}} \\
        \midrule
        \multicolumn{1}{l}{\begin{tabularx}{0.1\textwidth}{@{}X@{}}
            \textbf{Generated Solution After RLMEC}
        \end{tabularx}} & \multicolumn{1}{l}{\begin{tabularx}{0.8\textwidth}{@{}X@{}}
            Applying for a job can make someone feel stress even if they get it.\\Stress is an emotion.\\The answer is stress
        \end{tabularx}} \\
        \midrule
        \multicolumn{1}{l}{\begin{tabularx}{0.1\textwidth}{@{}X@{}}
            \textbf{Generated Solution After RFT}
        \end{tabularx}} & \multicolumn{1}{l}{\begin{tabularx}{0.8\textwidth}{@{}X@{}}
            Applying for a job is followed by attending an interview.\\Attending an interview gives \textcolor{red}{rejection or acceptance}.\\Rejection is an emotion.\\The answer is rejection
        \end{tabularx}} \\
        \midrule
        \multicolumn{1}{l}{\begin{tabularx}{0.1\textwidth}{@{}X@{}}
            \textbf{Generated Solution After DPO}
        \end{tabularx}} & \multicolumn{1}{l}{\begin{tabularx}{0.8\textwidth}{@{}X@{}}
            Frustration is the feeling of being upset or annoyed as a result of being \textcolor{red}{unable to change or achieve something}.\\Applying for a job can make someone feel frustration even if they get it.\\The answer is frustration
        \end{tabularx}} \\
        \midrule
        \multicolumn{1}{l}{\begin{tabularx}{0.1\textwidth}{@{}X@{}}
            \textbf{Generated Solution After PPO A2C}
        \end{tabularx}} & \multicolumn{1}{l}{\begin{tabularx}{0.8\textwidth}{@{}X@{}}
            Applying for a job is followed by attending an interview.\\Attending an interview gives an opportunity to sell one's skills and abilities to the employer.\\Employer decides the suitability of the candidate based on his skills and abilities.\\If the employer \textcolor{red}{rejects the candidate}, he feels frustrated.\\The answer is frustration
        \end{tabularx}} \\
        \bottomrule
    \end{tabular}
    \caption{The case study for question-answering tasks.}
    \label{case_study_qa}
\end{table*}

\begin{table*}[ht]
    \small
    \centering
    \begin{tabular}{ll}
        \toprule
        \multicolumn{1}{l}{\begin{tabularx}{0.1\textwidth}{@{}X@{}}
            \textbf{Problem}
        \end{tabularx}} & \multicolumn{1}{l}{\begin{tabularx}{0.8\textwidth}{@{}X@{}}
            Below is an instruction that describes a task. Write a response that appropriately completes the request.\\ \\ \#\#\# Instruction:\\A waiter had 12 customers. While 15 customers left he got 14 new customers. How many customers does he still have? \\ \\ \#\#\# Response:  Let's think step by step.
        \end{tabularx}} \\
        \midrule
        \multicolumn{1}{l}{\begin{tabularx}{0.1\textwidth}{@{}X@{}}
            \textbf{Generated Solution Before Training}
        \end{tabularx}} & \multicolumn{1}{l}{\begin{tabularx}{0.8\textwidth}{@{}X@{}}
            The waiter has \textcolor{red}{12-15=3} customers left.\\He got 14+3=17 customers.\\The answer is 17
        \end{tabularx}} \\
        \midrule
        \multicolumn{1}{l}{\begin{tabularx}{0.1\textwidth}{@{}X@{}}
            \textbf{Generated Solution After RLMEC}
        \end{tabularx}} & \multicolumn{1}{l}{\begin{tabularx}{0.8\textwidth}{@{}X@{}}
            The waiter had 12 customers + 14 customers = 26 customers.\\But since 15 customers left, he now has 26 customers - 15 customers = 11 customers.\\The answer is 11
        \end{tabularx}} \\
        \midrule
        \multicolumn{1}{l}{\begin{tabularx}{0.1\textwidth}{@{}X@{}}
            \textbf{Generated Solution After RFT}
        \end{tabularx}} & \multicolumn{1}{l}{\begin{tabularx}{0.8\textwidth}{@{}X@{}}
            The waiter has \textcolor{red}{12-15=3} customers left.\\He has 14+3=17 customers now.\\The answer is 17
        \end{tabularx}} \\
        \midrule
        \multicolumn{1}{l}{\begin{tabularx}{0.1\textwidth}{@{}X@{}}
            \textbf{Generated Solution After PPO A2C}
        \end{tabularx}} & \multicolumn{1}{l}{\begin{tabularx}{0.8\textwidth}{@{}X@{}}
            The waiter has \textcolor{red}{12-15=3} customers left.\\He got 14+3=17 customers.\\The answer is 17
        \end{tabularx}} \\
        \bottomrule
    \end{tabular}
    \caption{The case study for mathematical tasks.}
    \label{case_study_math}
\end{table*}

\section{Details for RLMEC}

\subsection{Prompts for Generative Reward Model Training}

We present the template of the prompt for teacher model distillation, and generative reward model training and inference in Table~\ref{prompt_tm} and Table~\ref{prompt_rm}, respectively.
In practice, the information (\ie question $q$, Ground-Truth Solution~$s$ and Generated Erroneous Solution~$\hat{s}$) should be filled into the corresponding curly brackets.
For error locating task, to better guide teacher model and generative reward model to figure out the first undesired step, we utilize the index to format the ground-truth solution.
The formatted solution is as follows,

[0] \texttt{The First Reasoning Step} $r_0$

[1] \texttt{The Second Reasoning Step} $r_1$

$\cdots$

[$n$] \texttt{The Last Reasoning Step} $r_n$

For the generative reward model, the training instruction and inference prompt are similar.
The target output of the training procedure (\ie the bold sentence in the table) will be removed during inference.

\subsection{Implementation Details for Experiments}
\label{implementation_details}
\paratitle{Datasets.}
We employ mathematical tasks and question-answering tasks for evaluation. 
Successfully solving these tasks necessitates LLMs to possess domain-specific knowledge and engage in systematic, step-by-step reasoning to reach the ultimate answer. 
The specifics of each dataset are delineated in Table~\ref{dataset}.

$\bullet$ \emph{Mathematical tasks} include GSM8k~\cite{gsm8k}, MATH~\cite{math}, SVAMP~\cite{svamp} and the mathematical problems in MMLU (MM)~\cite{mmlu_1,mmlu_2}. 
We adopt MathInstruct~\cite{mammoth} as the training set and eliminate the code samples. 
Given MathInstruct contains the training set of GSM8k and MATH, they are seen tasks for LLMs, while SVAMP and MM are unseen tasks.

$\bullet$ \emph{Question-answering tasks} contain ECQA~\cite{ecqa}, QASC~\cite{qasc}, OpenbookQA~\cite{openbookqa} and ARC-Easy~\cite{arc}.
We merge the training set of ECQA and QASC, and adopt the mixture as the training set in the experiment,
Therefore, ECQA and QASC are seen tasks for LLMs, while OpenbookQA and ARC are unseen tasks for LLMs.

\paratitle{Baselines.}
For a more comprehensive assessment, we incorporate three categories of methods as baseline approaches.
including supervised fine-tuning, alignment without reinforcement learning, and reinforcement learning.

$\bullet$ \emph{Supervised Fine-tuning} 
trains LLMs to imitate the human desired behavior. We conduct the SFT~\cite{instructgpt} and the Rejection sampling Fine-Tuning~(RFT)~\cite{rejection_sampling,rft} as the baseline methods.

$\bullet$ \emph{Alignment without Reinforcement Learning} is the method to align LLMs to human preference and prevent instability in reinforcement learning. Representative methods, \eg DPO~\cite{dpo}, CoH~\cite{coh}, and FIGA~\cite{figa} are conducted as the baseline.

$\bullet$ \emph{Reinforcement Learning} is the traditional method to guide LLMs to explore the world and learn from external feedback. 
PPO~\cite{ppo} is the classical algorithm to employ reinforcement learning.
We conduct the vanilla PPO~\cite{ppo} and Actor-Critic version of PPO (PPO A2C)~\cite{moss_rlhf} in the experiment.

Moreover, we also report the performance of base LLMs, including LLaMA 2~\cite{llama2}, Vicuna~\cite{vicuna}, and WizardLM~\cite{wizardlm}.

\paratitle{Hyper-Parameters Setting.}
In the experiment, we adopt Claude 2~\cite{claude2} as the teacher model.
For backbone LLMs, we utilize the mixture dataset of ECQA and QASC to fine-tune LLaMA 2~\cite{llama2} to obtain the domain-adapted SFT backbone model in QA tasks, and adopt MAmmoTH~\cite{mammoth} as the backbone model for mathematical tasks.
The backbone LLMs of the policy model and the generative reward model are the same SFT LLMs.
In the training procedure, we employ $5\times 10^{-6}$ as the learning rate for all tasks and train LLMs for 1 epoch.
Besides, we set 128 and 768 as the batch size for QA tasks and mathematical tasks.
For the value of $\varepsilon$, we leverage $0.3$ and $0.4$ for 7B model and 13B model, respectively.
Because the LLMs have adapted to the corresponding domain after training, we adopt the 0-shot setting during evaluation.

\section{Performance Analysis of RLMEC}
\label{analysis}

\subsection{Analysis of Generative Reward Model.}
The effectiveness of the generative reward model will influence the quality of the token-level rewards and the refined response.
Thus, we present the comparison of the teacher model and the generative reward model on QA tasks in Figure~\ref{tm_rm}.
We can observe that both the teacher model and the generative reward model can significantly reduce the edit distance and even perform slightly better than the teacher model.
That is because we utilize the two-stage prompting strategy to distillate knowledge from the teacher model and conduct the high-quality data to fine-tune the generative reward model.
Through fine-tuning, it can adapt to the erroneous solution rewriting task well.
Moreover, the teacher model and the generative reward model have shown similar performance on the accuracy of the refined responses, which verifies that the rewriting task can be easily learned by the LLMs with smaller parameters.
Besides, given the performance of RFT w/ TD and RFT w/ RD in Table~\ref{main_results}, we can observe that the higher accuracy of the refined responses will lead to higher performance in downstream tasks through simply supervised fine-tuning.

\subsection{Scaling Analysis of Reward Model.}
To explore the influence of the scale of the generative reward model, we conduct the experiment and present the results in Table~\ref{scale_of_rm}.
For both 7B and 13B LLMs, the rewriting model trained from the same backbone LLMs with the policy model performs better.
The potential reason might be that the policy model and the rewriting model with the same backbone model will have a similar distribution.
In this situation, the rewriting model can provide appropriate supervision signals and better guide the training process.

\subsection{Position of the First Error}
We conduct experiments about the position of the first error in the generated response after training.
The results are shown in Figure~\ref{error_shift}, respectively.
Rhe experiment on the position of the first error can    verify the effectiveness of RLMEC.
Compared with the backbone LLMs, the first error appears later after RLMEC.
For example, after RLMEC, the number of problems where the first error occurs before the final answer 7 steps (\ie the third column on the right) has increased, while the number of problems where the first error occurs more than 7 steps has decreased.
The reason is that LLMs focus on the mistakes and learn to correct the early errors during RLMEC.
In the ideal situation, all of the mistakes will be corrected through further training.
In contrast, after training through other methods, the position of the first error is irregular, which means that these methods do not consider the mistakes in the generated response and guide LLMs to learn to generate the correct solution without purposiveness.

\section{Case Study}
\label{case_study}

\subsection{Analysis of the Supervision Signals}
We present the case study about the reward from different methods in Table~\ref{reward}.
To better express the difference, we do not employ the clip mechanism in the case study.
From the results, we can observe that the reasoning step of the generated solution is correct but the final answer is error.
In PPO A2C, the reward will be calculated by the reward model and the critic model.
The tokens generated earlier will receive a lower reward, which is contradictory to reality.
That is because PPO A2C has assumed that the previous token will influence the last token.
In this case, once the generated solution contains the wrong answer, the rewards of the previous tokens are likely lower than the last tokens.
In contrast, we leverage the generative reward model to generate the reward in RLMEC.
The reward of the current token is calculated based on the previous tokens.
Therefore, the rewriting model in RLMEC can better indicate whether the token is correct and provide high-quality token-level supervision signals.
Besides, for the outcome-supervised method (\ie Vanilla PPO), the reward of each token is equivalent and is based on whether the generated solution is correct.
This method cannot describe the correctness of the tokens in the generated response.

\subsection{Analysis of the Generated Responses}

To further demonstrate the effectiveness of RLMEC, we present the case study about the performance of the LLMs trained by different methods in Table~\ref{case_study_qa} and Table~\ref{case_study_math}.
In both cases, our proposed RLMEC can help the LLMs to focus on the previous errors and correct the errors in the next time generation.
Concretely, in the question-answering tasks, the keywords of the problem are ``even if they get it''. 
After being trained through RLMEC, the LLMs can understand the meaning of the problem, figure the key point, and reach the correct answer.
However, through other methods, the LLM is still unable to grasp the key works in the problem and generate the answer about the emotion of losing the job.
Moreover, for mathematical problem, the LLM have made the mistake in calculating ``$12-15$''.
The LLM trained by baseline methods still make similar mistakes.
This case has shown that it is difficult for the previous methods to generate the supervised signals which can directly indicate the mistakes in the generated content and guide the LLMs to correct the errors.
In contrast, RLMEC leverages the generative reward model to provide the token-level supervised signals and guide the LLMs to focus on the mistakes.
Therefore, through RLMEC, the LLMs can correct the previous errors and obtain the correct answer.

\end{document}